\documentclass[conference]{IEEEtran}
\IEEEoverridecommandlockouts
\ifCLASSOPTIONcompsoc
    \usepackage[caption=false, font=normalsize, labelfont=sf, textfont=sf]{subfig}
\else
\usepackage[caption=false, font=footnotesize]{subfig}
\fi

\usepackage{url}
\usepackage{cite}
\usepackage{amsmath,amssymb,amsfonts}
\usepackage{algorithmic}
\usepackage{graphicx}
\usepackage{textcomp}
\usepackage{xcolor}
\usepackage{multirow}
\usepackage{lscape}
\usepackage{ragged2e}
\usepackage{booktabs}
\def\BibTeX{{\rm B\kern-.05em{\sc i\kern-.025em b}\kern-.08em
    T\kern-.1667em\lower.7ex\hbox{E}\kern-.125emX}}
\begin{document}

\title{Developing a Resource-Constraint EdgeAI model for Surface Defect Detection\\
}



\author{\IEEEauthorblockN{Atah Nuh Mih\IEEEauthorrefmark{1}, Hung Cao\IEEEauthorrefmark{1}, Asfia Kawnine\IEEEauthorrefmark{1}, Monica Wachowicz\IEEEauthorrefmark{1}\IEEEauthorrefmark{3}}

\IEEEauthorblockA{\IEEEauthorrefmark{1} \textit{Analytics Everywhere Lab, University of New Brunswick, Canada} \\
\IEEEauthorrefmark{3} \textit{RMIT University, Australia} \\\\}
}

\maketitle

\begin{abstract}
Resource constraints have restricted several EdgeAI applications to machine learning inference approaches, where models are trained on the cloud and deployed to the edge device. This poses challenges such as bandwidth, latency, and privacy associated with storing data off-site for model building. Training on the edge device can overcome these challenges by eliminating the need to transfer data to another device for storage and model development.  On-device training also provides robustness to data variations as models can be retrained on newly acquired data to improve performance. We therefore propose a lightweight EdgeAI architecture modified from Xception, for on-device training in a resource-constraint edge environment. We evaluate our model on a PCB defect detection task and compare its performance against existing lightweight models - MobileNetV2, EfficientNetV2B0, and MobileViT-XXS. The results of our experiment show that our model has a remarkable performance with a test accuracy of 73.45\% without pre-training. This is comparable to the test accuracy of non-pre-trained MobileViT-XXS (75.40\%) and much better than other non-pre-trained models (MobileNetV2 - 50.05\%, EfficientNetV2B0 - 54.30\%). The test accuracy of our model without pre-training is comparable to pre-trained MobileNetV2 model - 75.45\% and better than pre-trained EfficientNetV2B0 model - 58.10\%. In terms of memory efficiency, our model performs better than EfficientNetV2B0 and MobileViT-XXS. We find that the resource efficiency of machine learning models does not solely depend on the number of parameters but also depends on architectural considerations. Our method can be applied to other resource-constraint applications while maintaining significant performance. 

\end{abstract}


\section{Introduction}




Recent advances in Artificial Intelligence (AI) and Machine Learning (ML) boost the proliferation of many AI-based applications and services. It is undeniable that new AI models such as large language models \cite{zhao2023survey} and diffusion models \cite{croitoru2023diffusion} are changing our lifestyles. However, these AI models require intensive computational resources such as CPU, GPU, memory, and network that only the cloud can offer. Cloud computing has been a key enabler of many new technologies, such as IoT, and AR/VR, by providing virtually unlimited resources, including on-demand storage and high computing power. By leveraging these advantages, many powerful AI models such as Segment Everything \cite{kirillov2023segment} are trained and deployed in the cloud. However, there are several drawbacks when implementing AI models that completely rely on the cloud. For example, the AI models running on the cloud will depend totally on the external infrastructure, leading to potential service interruptions and downtime if the cloud service provider experiences outages or technical issues. Moreover, cloud-based AI models may face latency and performance issues since their quality of services is closely tied to network quality. Unstable connectivity, bandwidth limitations, and network delays could cause many AI service interruptions.

These challenges motivate the need to rely on edge computing for training ML models. The rapid development of mobile chipsets and hardware accelerators has improved edge devices' computing power significantly \cite{shi2020communication}. This has led to a shift from deploying AI models in the cloud to the edge, where AI functionalities are diffused, converged, and embedded into resource-constrained devices in physical proximity to the users, such as micro data centers, cloudlets, edge nodes, routers, and smart gateways. However, this shift wave is only partially implemented and has not fully taken advantage of the power of edge computing. The literature evidences this since existing solutions only deploy the inference AI models at the edge \cite{sallang2021cnn} \cite{tsukada2020neural}.

Training on the edge can prove beneficial in terms of variations between training and deployment environments and also address the viewpoint problem \cite{kukreja2019training}. However, the main challenge of training on the edge is the availability of computing resources, as modern deep learning architectures are designed to be computationally intensive. Although various lightweight deep learning models \cite{tan2021efficientnetv2} \cite{mehta2021mobilevit} \cite{sandler2018mobilenetv2} have been proposed, they do not perform as well as their heavyweight counterparts. This leaves a research gap in developing deep learning architectures suitable for training on resource-constraint edge devices. We therefore aim to answer the following research question: \textit{``Can we build deep learning models, train and deploy them directly at the edge with limited resources while maintaining acceptable performance?"}


In this paper, we highlight the importance of aligning research towards the development of deep learning techniques designed for edge devices whose performance is comparable to state-of-the-art deep learning models. We present a PCB defect detection task using the approach in \cite{nuh2023transferd2} on an edge device. We provide a modification of the Xception architecture that makes it suitable for training on the edge device. We experiment with different existing lightweight models at the edge, evaluate their performance on defect detection, and compare them to our model. 




The main contributions of this work are as follows:
\begin{enumerate}
    \item We present a novel lightweight architecture based on Xception, which facilitates on-device training in a resource-constraint environment. 
    \item We evaluate the performance of our model in terms of accuracy, memory usage, GPU utilization, and power consumption on a PCB defect detection task, and compare the performance against existing lightweight architectures. 
    \item We explore the benefit of using transfer learning on lightweight architectures in terms of accuracy and resource efficiency of the models.  
\end{enumerate}


\section{Related work}
\subsection{AI models at the Edge}
Many studies have implemented artificial intelligence on edge. Nikouei et al. \cite{nikouei2018real} developed a lightweight CNN (L-CNN) using depthwise separable convolution and a Single Shot Multi-Box Detector (SSD) for human object detection and deployed the model on an edge device, Sallang et al \cite{sallang2021cnn} deployed a MobileNetV2-based SSD on a Raspberry Pi 4 for smart waste management; and Sreekumar et al \cite{sreekumar2018real} designed a real-time traffic pattern collection method using YOLOv2 deployed on an edge device.

Beyond the use of edge devices for deploying machine learning models, other authors explored performing on-device training. Kukreja et al. \cite{kukreja2019training} proposed using a student-teacher model for training, where a teacher model is trained on an object and used to update the dataset with different viewpoints on which student models are trained. They also discuss the use of checkpointing to reduce the memory consumption of the training process. Tsukada et al. \cite{tsukada2020neural} proposed an On-device Learning Anomaly Detector (ONLAD), which combines sequential learning with semi-supervision and an autoencoder to reduce computational cost. They developed a hardware implementation of their method called ONLAD Core, on which they performed on-device training. 

The absence of extensive studies on performing on-device training leaves a research gap to be explored. We leverage this gap by taking a novel approach to on-device training, specifically in defect detection. We modify an existing deep neural network to make it lightweight and computationally efficient enough to train on the edge. We then perform inference on the same device itself, eliminating the overhead of deploying the model to a different device for inference.

\subsection{Defect Detection}
Fault detection is a prevalent topic in smart manufacturing, with many works that explore various methods to solve this challenge. He et al. \cite{he2019end} proposed a steel plate defect detection approach using a baseline convolution neural network and a multilevel feature fusion network for the localisation of defects. Ding et al \cite{ding2019tdd} proposed Tiny Defect Detection for identifying printed circuit board (PCB defects) using Faster R-CNN. Zhang et al \cite{zhang2018improved} also proposed a PCB defect detection approach using a pre-trained VGG model for feature extraction and a sliding window for identifying the defects. 

Several Edge AI solutions have been explored for detecting imperfections. Zhu et al \cite{zhuZ2020modified} proposed a lightweight modification of DenseNet, which they deployed on an edge camera for fabric defect detection in an industrial setting. Song et al \cite{song2021efficientdet} also designed a fabric defect detection framework based on EfficientDet - a scalable and efficient lightweight object detection model. They trained the model on a workstation and deployed it on an Nvidia Jetson TX2 for inference.   

To the best of our knowledge, widespread research has not been conducted towards implementing a PCB defect detection task, which involves training and deploying on the edge device. We, therefore, explore this research gap in our paper.

\section{Proposed Method}




\subsection{Overview of Approach}
\begin{figure*}[t!] 
    \centering
  \subfloat[Overview Approach\label{fig_overview}]{%
       \includegraphics[width=0.48\linewidth]{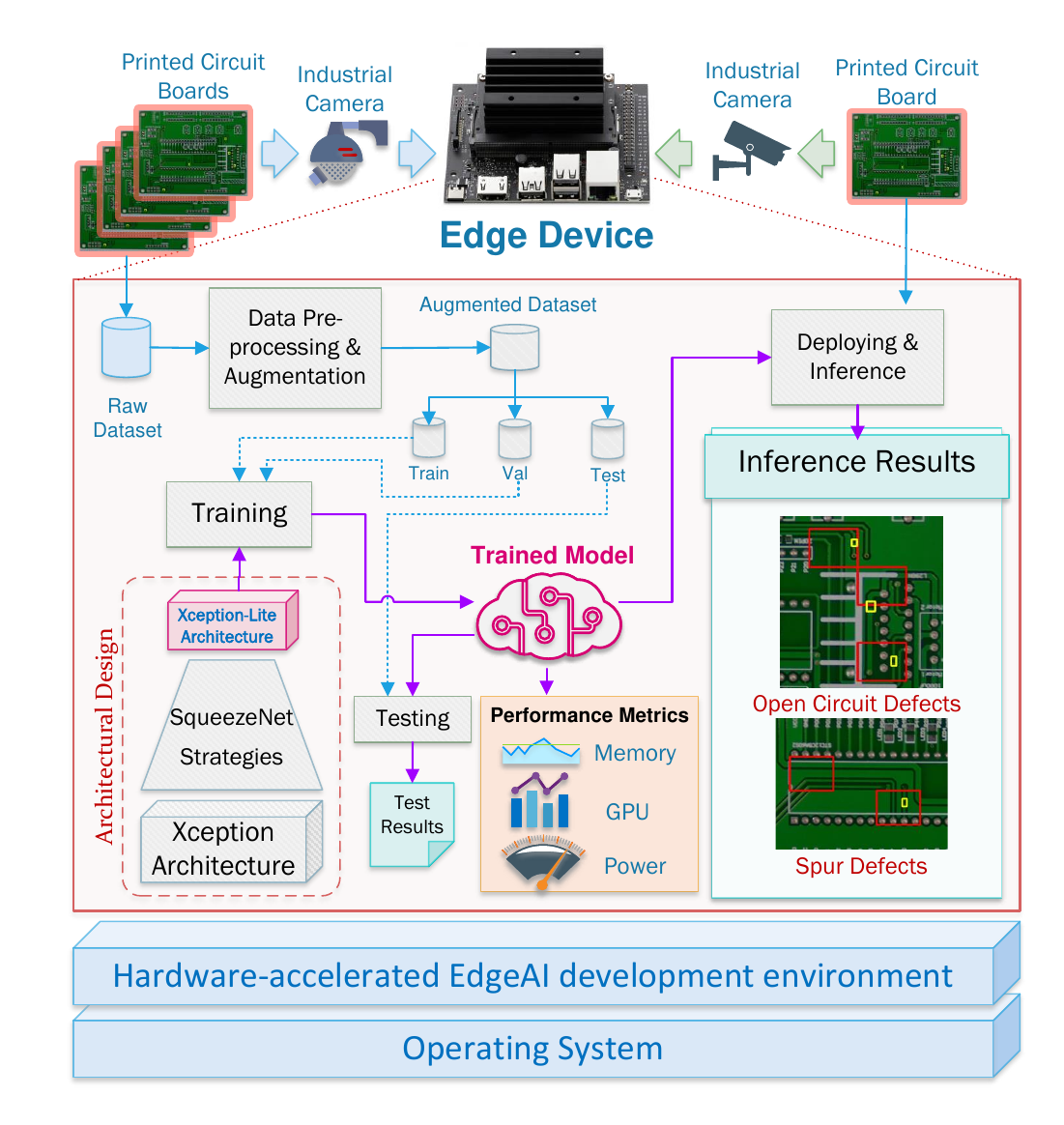}}
    \hfill
  \subfloat[Proposed Architecture (adapted and modified from \cite{chollet2017xception})\label{fig_xception-lite_arch}]{%
        \includegraphics[width=0.5\linewidth]{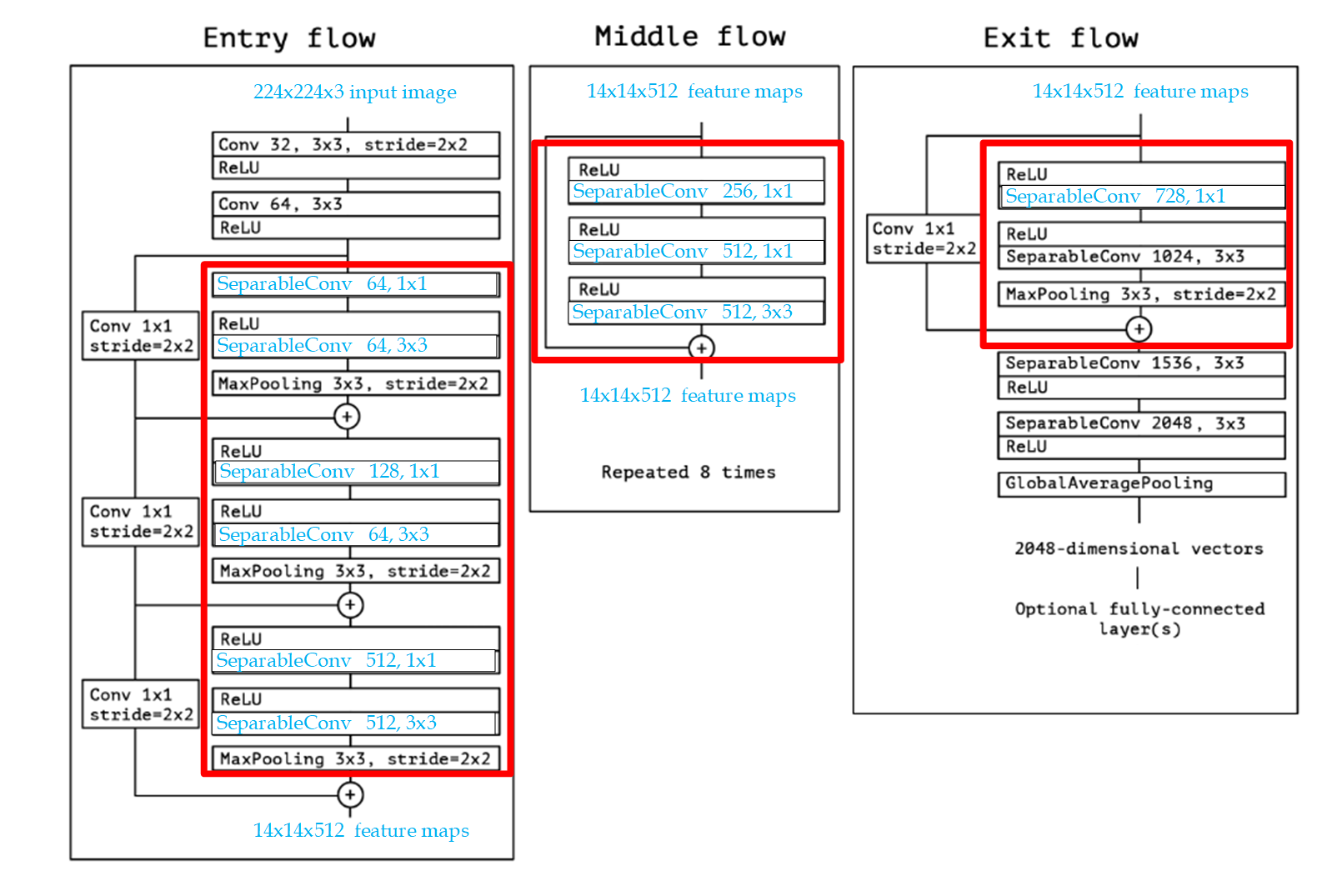}}    
  \caption{The proposed method to build a lightweight EdgeAI model.}
        \label{fig:methodology}
\end{figure*}

We present a PCB defect detection scenario in an industrial setting as shown in Figure \ref{fig_overview}. The simulated setup consists of an edge device with a hardware-accelerated EdgeAI development environment on which machine learning training and deployment are conducted.

A camera captures images of manufactured printed circuit boards and transfers them to an edge device, where they are stored as a raw dataset. This dataset is pre-processed and augmented using the technique described in \cite{nuh2023transferd2}. 
We split the dataset into the training, validation, and testing sets.

We build our proposed architecture by modifying the Xception architecture\cite{chollet2017xception} using SqueezeNet strategies \cite{iandola2016squeezenet}. We train the model on the edge device using the training and validation sets of the augmented dataset and evaluate its performance on the test set. We also measure the performance of the model in terms of memory usage, GPU utilisation, and power consumption. The trained model is then deployed on-device for inference on newly captured images. The model makes predictions for new images to identify areas containing defects and highlights them with a bounding box.  

We discuss in detail, the design of our architecture in the next section.

\subsection{Xception}
The Xception architecture \cite{chollet2017xception} consists of a linear stack of depthwise separable convolutions. It has 36 convolutional layers structured into 14 modules with residual connections around them, except for the first and last modules. These modules are divided into an Entry flow, a Middle flow, and an Exit flow.  


The modular nature of this architecture makes it easy to modify and re-design. Depthwise separable convolution and residual connections are two attributes that have been reported to be essential in the design of lighter and more efficient network architectures \cite{wangX2020convergence} \cite{nikouei2018real} \cite{xuR2022efficient}. We, therefore, postulate that other design factors can be considered to reduce the size of the network.

\subsection{SqueezeNet}\label{meth:squeezenet}
Iandola et al \cite{iandola2016squeezenet}, proposed SqueezeNet as a lightweight CNN model with fewer parameters but with comparable performance to a heavyweight model. They describe three strategies to reduce the number of parameters of a CNN architecture: Strategy (1) Replacing 3x3 filters with 1x1 filters; Strategy (2) Decreasing the number of input channels to 3x3 filters; and Strategy (3) Downsampling late in the network so that convolutional layers have large activation maps.

The authors also present a \textit{fire module} consisting of a \textit{squeeze} convolutional layer (containing only 1x1 filters), followed by two \textit{expand} layers (a mix of 1x1 and 3x3 filters). Using these design considerations, they propose a SqueezeNet architecture containing a single convolutional layer, followed by 8 \textit{fire modules}, and a final convolutional layer.   
SqueezeNet showed a similar performance to AlexNet in the ImageNet classification task, with a Top-5 accuracy of 80.3\%, while being 50x lighter.

\subsection{Squeezing Xception}
Xception, in itself, is not a lightweight architecture and it was impossible to implement it on the edge device. Therefore, we propose a variation of the Xception architecture that aims to reduce its size while maintaining high accuracy in the scope of this experiment. We modify the Xception network to include Strategies (1) and (2) of SqueezeNet, which aim to reduce the number of parameters while preserving accuracy. We maintain the original macro-architectural design of Xception (i.e. Entry flow, Middle flow, and Exit flow), but alter the micro-architecture (i.e. varying the number of filters and channels). 

We begin by replacing all first 3x3 filters in Separable Convolution layers with 1x1 filters, thereby satisfying Strategy (1). This technique reduces the number of filters nine-fold and subsequently reduces the number of parameters.    

The number of parameters of a layer is defined by: 
\begin{equation}
    \omega = N_{channels} * M_{filters} * \Psi_{filter} 
\end{equation}

where \\ 
    $\omega$ is the number of parameters, \\
    $N_{channels}$ is the number of channels \\
    $M_{filters}$ is the number of filters \\
    $\Psi_{filter}$ is the dimensions of the filter, e.g. 1x1 or 3x3
    
Strategy (1) already explores reducing the number of filters by replacing 3x3 filters with 1x1 filters. By implementing Strategy (1), we have a fixed number of filters with fixed filter sizes. Equation 1 becomes a linear relationship between the number of parameters and the number of channels. This relationship is exploited by Strategy (2) which proposes decreasing the number of input channels into a layer. 

We reduce the number of channels in the Entry Flow and Middle Flow to ensure that fewer channels are fed into subsequent layers of the network. The \textit{fire module} described in Section \ref{meth:squeezenet} also provides an implementation of Strategy (2). The \textit{squeeze} layer of the \textit{fire module} reduces the number of channels that are fed into the final \textit{expand} layer (3x3 filters).  

To implement a \textit{fire module}, we replace the first layer of the Middle Flow modules with a layer containing 1x1 filters as the \textit{squeeze} layer. The next layer becomes the first \textit{expand} layer, and we also replace its 3x3 filters with 1x1 filters. The third layer then becomes the final \textit{expand} layer with 3x3 filters. 

A summary of the network is shown in Figure \ref{fig_xception-lite_arch}.


\section{Implementation}
\subsection{System Specifications}
We conduct our study using an A203 Mini PC built with Nvidia's Jetson Xavier NX 8GB module, 128GB SSD, and a pre-installed JetPack 5.0.2 on Ubuntu 20.04. The A203 Mini PC is a powerful edge computer that brings AI to the edge. With up to 21 TOPS and an integrated GPU, it provides AI computational capabilities for smart cities, industrial automation, and smart manufacturing.
The device has several power modes that deliver different levels of performance. We set our device to the maximum power settings \textit{MODE 20W 6CORE}.


\subsection{Dataset}\label{dataset}
We use the PCB Defect Dataset \cite{huang2019pcb}, which contains 1386 images with 6 types of defetcs (Missing Hole, Mouse Bite, Open Circuit, Short, Spur, Spurious Copper). The method in \cite{nuh2023transferd2} was successful in evaluating the defect detection on new classes of images not included in the training set. We therefore separate the Open Circuit and Spur classes from the rest of the classes to later evaluate performance on unseen classes of defects. 

The authors also describe a pre-processing technique to obtain a large dataset from a smaller number of images. The dataset generated occupies a much smaller disk space compared to the original dataset, making it convenient for devices with low storage capacity. The technique generates a new dataset of just two classes from the original dataset: 0 (for non-defective images) and 1 (for defective images). 

We follow this approach using the rest of the classes (Missing Hole, Mouse Bite, Short, and Spurious Copper). Images from the original PCB dataset are split into tiles and assigned to the new classes in the generated dataset depending on whether they contain a defect or not. We generate a new dataset of 20,000 images and split it into the training, validation, and testing datasets in the ratio 7:2:1.




\subsection{Building the Models} \label{model_design}
We implement our proposed lightweight model as shown in Fig \ref{fig_overview}. The input images are resized to 224x224 for uniformity before feeding them into the model's input layer. We build a classification head consisting of a global average pooling layer, a dropout layer, and an output layer with a sigmoid activation function on top of the architecture. We also experiment with other lightweight models - MobileNetV2, EfficientNetV2B0, and MobileViT-XXS; while maintaining the classification head. 
We use a batch size of 16 across the models and train for 60 epochs. We report the results in Section \ref{exp_results}


\section{Results and Discussions} \label{section:results}
\subsection{Experimental Results}\label{exp_results}
\begin{figure}[h!] 
    \centering
    \subfloat[Training Accuracy\label{fig_trainAccuracy}]{%
    \includegraphics[width=0.45\linewidth]{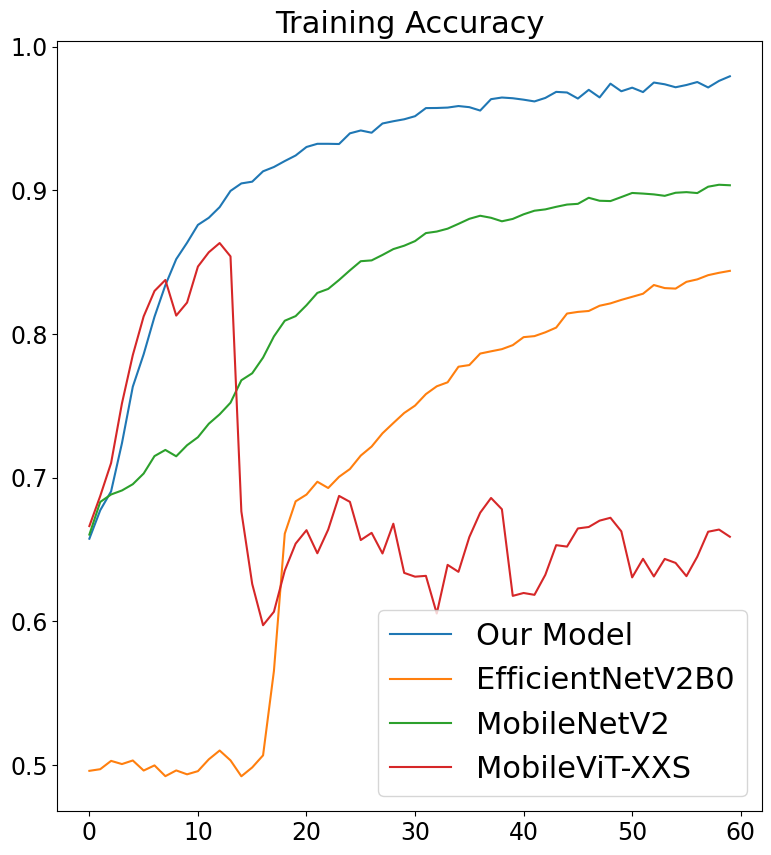}}
  \subfloat[Training Loss\label{fig_trainLoss}]{%
        \includegraphics[width=0.45\linewidth]{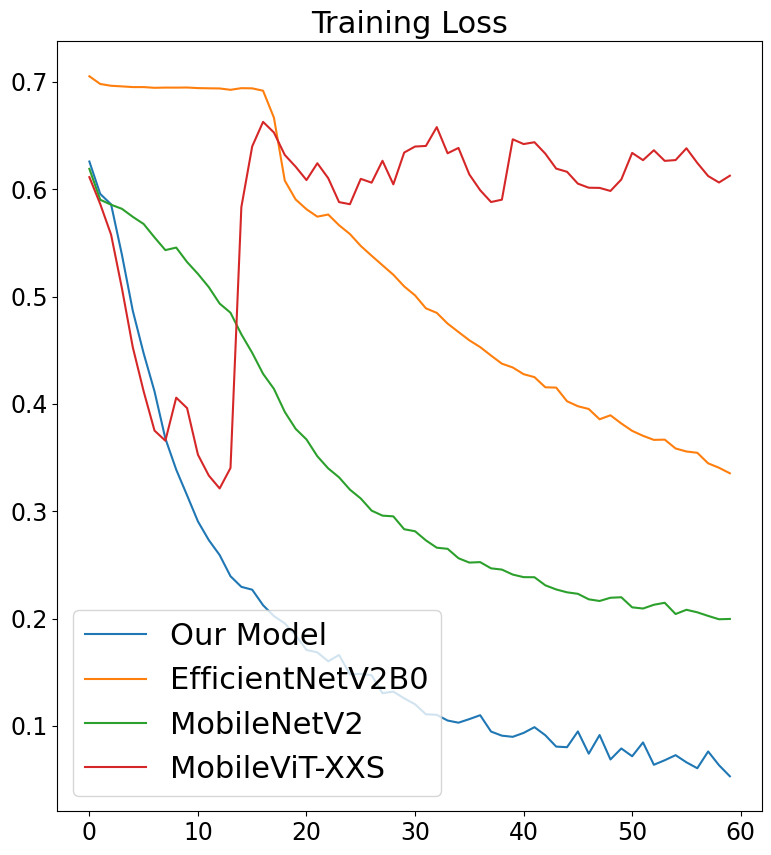}}    
  \caption{Comparison of Training Accuracy and Training Loss}
        \label{fig:train_results}
\end{figure}

We report the findings of our experiments in this section. We first evaluate the training accuracy and loss of the various models, and present the results in Figure \ref{fig:train_results}. We then provide a general comparison of the models' performance on the edge device in Table \ref{tab:models_compare}. We exclude a comparison for the default Xception as its training exceeded the edge device's capacity. Similarly, SqueezeNet exceeded the device's GPU allocation and its results are not included. 

The training results in Figure \ref{fig:train_results} show that our proposed architecture has the best training performance among the four models. It converges faster than the other models and has a better final accuracy. MobileNetV2 has the second best performance among the models. EfficientNetV2B0 has a poor performance at the start of training, but its performance sharply improves at later epochs. MobileViT-XXS has a high performance at the start of training (comparable to our proposed model). This drops sharply in subsequent epochs, resulting in the worst training performance. 

\begin{table*}[h!]
    \centering
    \caption{Comparison of model performance on the edge}
    \begin{tabular}{ccccccccc}
\hline\hline
\textbf{Model} & \textbf{Configurations} &\textbf{Parameters}  & \textbf{Train Acc} &\textbf{Test Acc} & \textbf{Avg Time/epoch} & \textbf{Avg Mem Used} & \textbf{Avg GPU Used} & \textbf{Avg Pow Cons}\\
\hline
\textcolor{blue}{Proposed Model} & \textcolor{blue}{no pre-training} & \textcolor{blue}{11.2M} & \textcolor{red}{\textbf{97.93\%}} & \textcolor{red}{\textbf{73.45\%}} & \textcolor{blue}{564.94s} & \textcolor{blue}{0.9074 GB} & \textcolor{blue}{92.84\%} & \textcolor{blue}{18.07W}\\\hline
EfficientNetV2B0 & no pre-training & 5.9M & 84.39\% & 54.30\% & 357.94s & 0.9339 GB & \textcolor{red}{\textbf{85.12\%}} & 16.42W\\
MobileNetV2 & no pre-training & 2.2M & 90.35\% & 50.05\% & \textcolor{red}{\textbf{270.09s}} & \textcolor{red}{\textbf{0.8838 GB}} & 87.14\% & \textcolor{red}{\textbf{15.19W}}\\
MobileViT-XXS & no pre-training & 1.3M & 65.89\% & 67.25\% & 693.81s & 0.9303 GB & 89.77\% & 15.90W\\
\hline
\end{tabular}

   
    \label{tab:models_compare}
\end{table*}

From Table \ref{tab:models_compare}, we observe that MobileNetV2 has a training accuracy of 90.35\% and a test accuracy of 50.05\%; EfficientNetV2B0 has a training accuracy of 84.39\% and a test accuracy of 54.30\%. These results show that MobileNetV2 and EfficientNetV2B0 overfit to the training data. MobileViT-XXS has a training accuracy of 65.89\% and a similar test accuracy of 67.25\%, which indicates a stable performance on both training and testing. Our model has the best performance on the training data (97.93\%) and the best test performance (73.45\%) among the models. 


We compare the test accuracies of the models and their number of parameters. EfficientNetV2B0 (5.9M) and MobileNetV2 (2.2M) with few parameters have very low test accuracies (50.05\% and 54.30\% respectively). MobileViT-XXS (1.3M) with fewer parameters has a better test accuracy (67.25\%) than MobileNetV2 and EfficientNetV2B0. Our architecture, with the most number of parameters (11.2M), has the best test score (73.45\%). The lack of a direct relationship between the number of parameters in EfficientNetV2B0, MobileNetV2, and MobileViT-XXS leads us to conclude that the number of parameters does not reflect the accuracy of the model. 



\begin{figure*}[h!] 
    \centering
  \subfloat[Memory Usage Pattern During Training\label{fig:res_mem_use}]{%
       \includegraphics[width=0.3\linewidth]{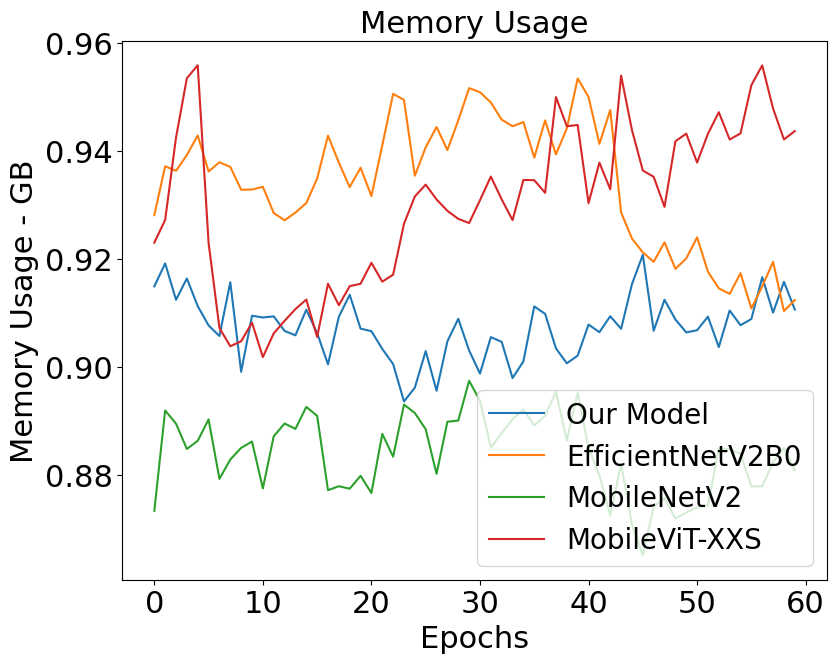}}
    \hfill
  \subfloat[GPU Usage During Training\label{fig:res_gpu_use}]{%
        \includegraphics[width=0.3\linewidth]{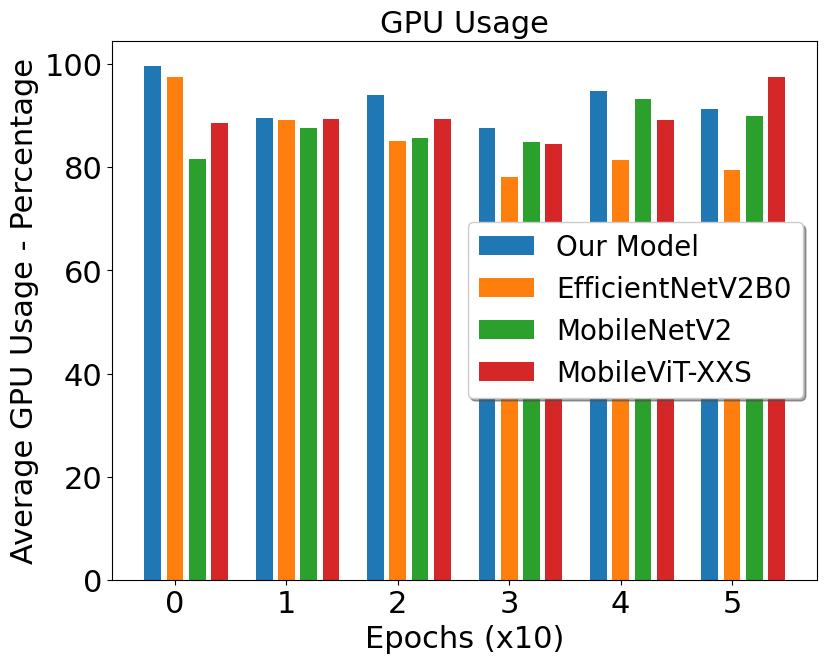}}
    \hfill
  \subfloat[Power Consumption During Training\label{fig:res_pow_cons}]{%
        \includegraphics[width=0.3\linewidth]{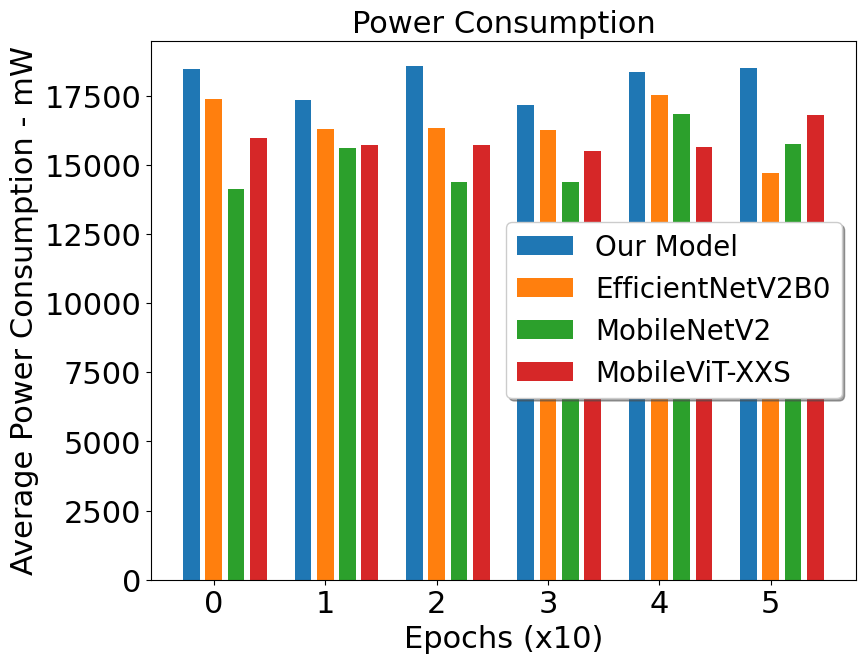}}
  \caption{Resource Consumption Pattern}
  \label{fig:detection_results}
\end{figure*}

\subsection{Performance Benchmark}
\subsubsection{Memory Usage}
We present the results of memory usage during training for the models in Figure \ref{fig:res_mem_use}.
We observe that MobileNetV2 has the lowest memory usage pattern throughout training, and our proposed architecture has the second lowest memory usage pattern. MobileViT-XXS has an initially high memory usage at the start of training, which steeply decreases and then rises again throughout training. EfficientNetV2B0 has the highest memory usage during most of the training process, but sharply drops towards the end of the training. 

We infer that our model is very memory efficient although it has the most number of parameters. It is only surpassed by MobileNetV2 and it has a better memory usage pattern than the much lighter EfficientNetV2B0 and MobileViT-XXS.  

 
\subsubsection{GPU Usage}
We report the results of GPU utilisation in Figure \ref{fig:res_gpu_use}.Our architecture has the highest average GPU usage during most of the training intervals. EfficientNetV2B0 has a high GPU utilisation at the start of training, but as training proceeds, the usage significantly drops below that of other models. MobileNetV2 has a low GPU usage during the early training epochs, but the usage increases towards the end of training. MobileViT-XXS has a consistent GPU utilization during most of the training, but at the end of training, its GPU usage surpasses all the models.

\subsubsection{Power Consumption}
Fig \ref{fig:res_pow_cons} shows the power consumption pattern during various intervals of training. We observe that our architecture has the highest power consumption through the entire training process. EfficientNetV2B0 has the second highest power consumption throughout most of the training duration, but towards the end of training, its power consumption drops below all other models. MobileNetV2 has the least power consumption for most of the epochs, but this increases towards the end of training. MobileVit-XXS has a consistent power consumption throughout most of the training duration.

\subsection{Defect Detection}\label{results_detection}
Nuh Mih et al. \cite{nuh2023transferd2} proposed a simple defect detection approach, which involves splitting an image into a 10x10 grid and passing each tile into the model for prediction. The prediction results are then used to identify defective areas in the original full image using pseudo bounding boxes. We use this method for defect detection on images from the two held-out classes (Open Circuit and Spur) of the original PCB Defect Dataset to evaluate models' performance on unseen classes of data.

For comparison purposes, we replicate the experiment in \cite{nuh2023transferd2} using transfer learning with Xception on an Intel(R) Core(TM) i7-4790 CPU desktop with 16GB memory, and present the results as our benchmark. We show the results in Figure \ref{fig:detection_results_benchmark} (see Appendix). The yellow bounding boxes represent the true annotations, while the red bounding boxes represent the tiles predicted as defective by the model. These results show the success of the approach on the PCB defect detection with unseen classes of defects.
We evaluate the performance of the models on the defect detection task and show the results in Figure \ref{fig:detection_results_OC} (see Appendix). We observe that EfficientNetV2B0 makes false positives predictions on both classes of defects, while MobileNetV2 fails to predict any defective tile (all false negatives). MobileViT-XXS has only one prediction on the Open Circuit class (false positive) and no prediction on the Spur class. Our proposed model fails to make a clear differentiation between defective and non-defective tiles resulting in several false positives. As the defects are small, the model learns general features of the image such as the circuitry, which are common in both defective and non-defective classes. The model then recognizes these features in both classes of images and therefore results in false positives shown in the results. 

In contrast, the benchmark Xception with transfer learning (Figure \ref{fig:detection_results_benchmark} in the Appendix) correctly classifies the defective regions of the image, with only a few false positives. The performance improvement is due to the fact that this model was pre-trained on ImageNet. We believe that pre-training our model on ImageNet would yield similar results, however, we leave this for future works.

\begin{table*}[h!]
    \centering
    \caption{Comparison of Pre-trained Models vs Our Model with no pre-training}
    \begin{tabular}{cccccccc}
\hline\hline
\textbf{Model} & \textbf{Configuration} & \textbf{Train Acc} & \textbf{Test Acc} & \textbf{Avg Time/epoch} & \textbf{Avg Mem Used} & \textbf{Avg GPU Used} & \textbf{Avg Pow Cons}\\
\hline
EfficientNetV2B0 & pre-trained & 56.28\% & 58.10\% & 134.94s & \textbf{0.8767GB} & \textbf{79.87\%} & \textbf{14.78W}\\
MobileNetV2 & pre-trained & 74.68\% & \textbf{75.45\%} & \textbf{107.34s} & 0.8795GB & 80.37\% & 15.21W\\

MobileViT-XXS & not available & - & - & - & - & - & -\\
\hline
\textcolor{blue}{Proposed Model} & \textcolor{blue}{no pre-training} & \textcolor{blue}{97.93\%} & \textcolor{blue}{73.45\%} & \textcolor{blue}{564.94s} & \textcolor{blue}{0.9074GB} & \textcolor{blue}{92.84\%} & \textcolor{blue}{18.07W} \\
\hline
\end{tabular}
   
    \label{tab:TL_compare}
\end{table*}

\subsection{Our Proposed Model vs Pre-trained Models} \label{TL_results}
We implement the defect detection method using MobileNetV2 and EfficientNetV2B0 pre-trained on the ImageNet dataset. We evaluate their performance and present our findings in Table \ref{tab:TL_compare}. We provide this separate analysis of the results from the models without pre-trained weights as their implementation details are not the same. However, we include the results of our proposed model for our discussion, but we do not include results for MobileViT-XXS in this comparison as its pre-trained model is not available on the Keras library.


From Table \ref{tab:TL_compare}, we observe that our model has a better training accuracy (97.93\%) than the pre-trained models - MobileNetV2 (74.68\%) and EfficientNetV2B0 (56.28\%). However, on the test set, MobileNetV2 has the best performance (75.45\%) as compared to EfficientNetV2B0 (58.10\%) and our model (73.45\%). Pre-trained EfficientNetV2B0 has the best performance in terms of memory usage (0.8767GB), GPU usage (79.87\%), and power consumption (14.78W). 

By comparing Table \ref{tab:TL_compare} with Table \ref{tab:models_compare}, we observe that the performance of the models improve significantly when using pre-trained weights, thereby pointing out the benefits of using transfer learning. The pre-trained models also have stable performance on training and testing with similar accuracies in both tasks. 

This comparison also shows improvements in terms of resource consumption as the pre-trained models now consume less memory, GPU, and power than when trained without transfer learning. The performance gain is significant with EfficientNetV2B0, which now outperforms MobileNetV2 on all resource consumption metrics. MobileNetV2 only has a faster training time than EfficientNetV2B0. 


We also evaluate the performance of these models on the defect detection task described in \ref{results_detection} and present our findings in Figure \ref{fig:detection_results_TL} (see Appendix).

    

We now observe improvements in the defect detection with MobileNetV2 as it makes more predictions than before (non-pretrained model). EfficientNetV2B0 on the other hand, does not show any improvement. 

\section{Discussion and Conclusion}


From the results discussed in Section \ref{exp_results}, we infer that two of the existing lightweight models fail to differentiate between defective and non-defective tiles, resulting in a low test accuracy $\sim 50\%$. These models benefit from the millions of images available in benchmark datasets on which they are trained to achieve good performance. They leverage large data, adequate training times, and sufficient computing resources at their disposal to circumvent their shortcomings. However, on edge devices where these factors are unavailable, the models fail to provide any significant performance.  


In terms of resource consumption, we observe that the size of the model is not the only factor affecting resource efficiency of the models. For example, EfficientNetV2B0 has the highest average memory usage, but the least average GPU utilisation. MobileViT-XXS has the highest average time per epoch while having the least number of parameters. While MobileNetV2 does not have the least number of parameters, it is the most resource-efficient model in our evaluation. Our model has a better training time and memory usage than MobileViT-XXS, despite having 10x more parameters than MobileViT-XXS. 

We conclude that architectural design considerations also play a major role in resource consumption. For example, the transformer in the MobileViT-XXS architecture suffers from decrease in performance on mobile devices due to the lack of device-level optimisations that improve latency and memory access \cite{mehta2021mobilevit}. EfficientNetV2B0 \cite{tan2021efficientnetv2} combines network design and progressive training (adaptively adjusting the regularization and image sizes for improved performance on its evaluation task). The lack of these optimizations result in poor performance when applied to a different task.


To obtain a better performance in on-device training, lightweight models can benefit from transfer learning. The results in Table \ref{tab:TL_compare} show that the pre-trained MobileNetV2 performs better than our proposed model by just 2\%. MobileNetV2 benefits from initialized weights trained on a much larger ImageNet dataset to obtain improved performance. Meanwhile our model, which begins with random weight initialisation, almost has the same performance as the pre-trained MobileNetV2 and outperforms the pre-trained EfficientNetV2B0. The benefits of initialized weights are not only limited to accuracy improvements, but also result in better resource consumption. Pre-trained models converge faster and are therefore computationally less intensive.  

We postulate that if our model was pre-trained on a larger dataset such as ImageNet, its performance would significantly improve as in the case of pre-trained MobileNetV2 and pre-trained EfficientNetV2B0. However, this justification can be proven in future works. 

Although our model was not optimal on resource consumption metrics, we believe more techniques can be explored to further reduce the size of the model and obtain better performance in this aspect. This can also enable us to optimize the model for training on edge devices with various computing capabilities. We leave this consideration for future work.

\bibliographystyle{ieeetr}
\bibliography{ref.bib}
\section{Appendix}
\renewcommand{\thefigure}{A.\arabic{figure}}
\setcounter{figure}{0} 

\begin{figure}[h!]
    \centering
  \subfloat[MobileNetV2: Open Circuit\label{3a}]{%
       \includegraphics[width=0.49\linewidth]{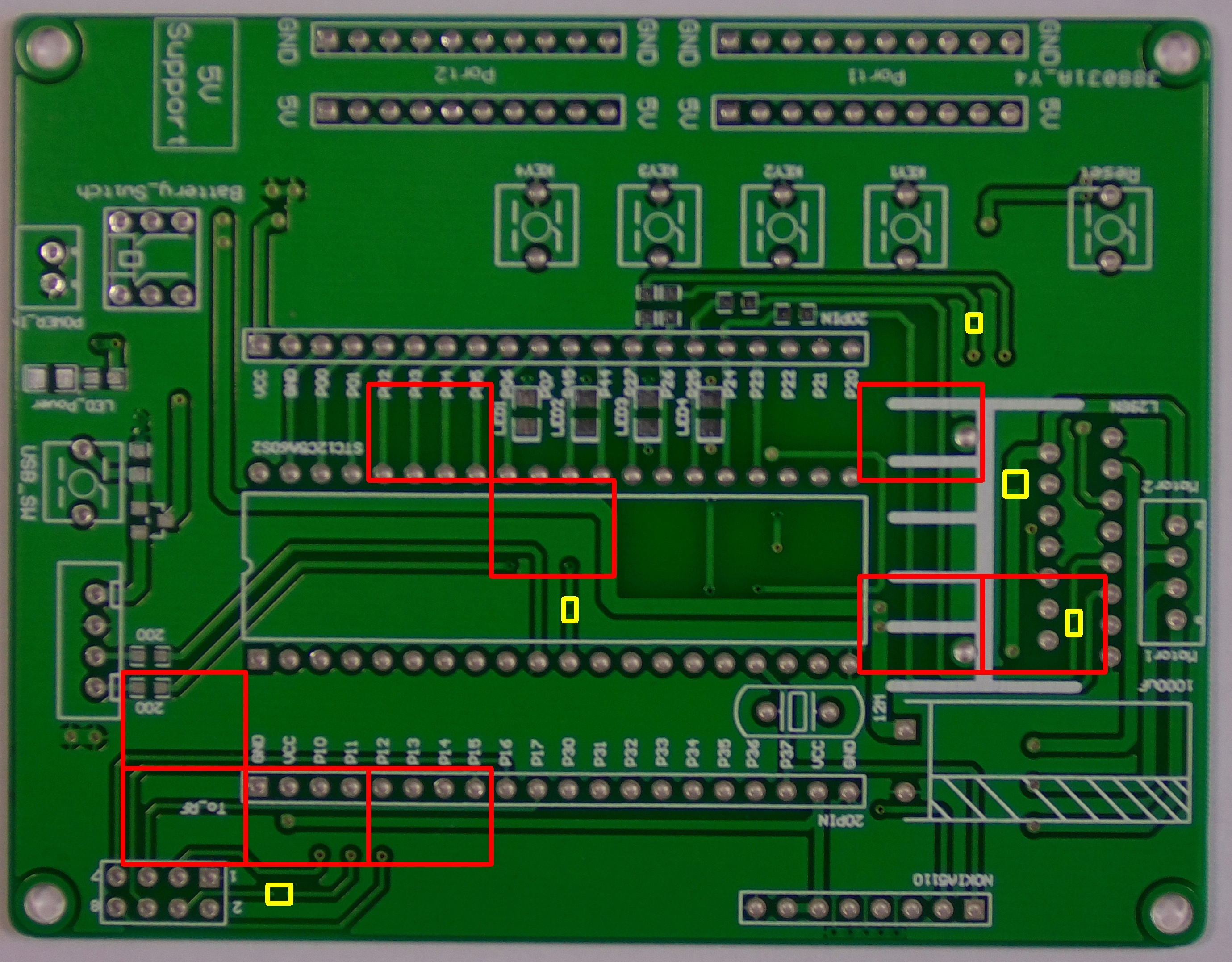}}
    \hfill
  \subfloat[MobileNetV2: Spur\label{3a}]{%
       \includegraphics[width=0.48\linewidth]{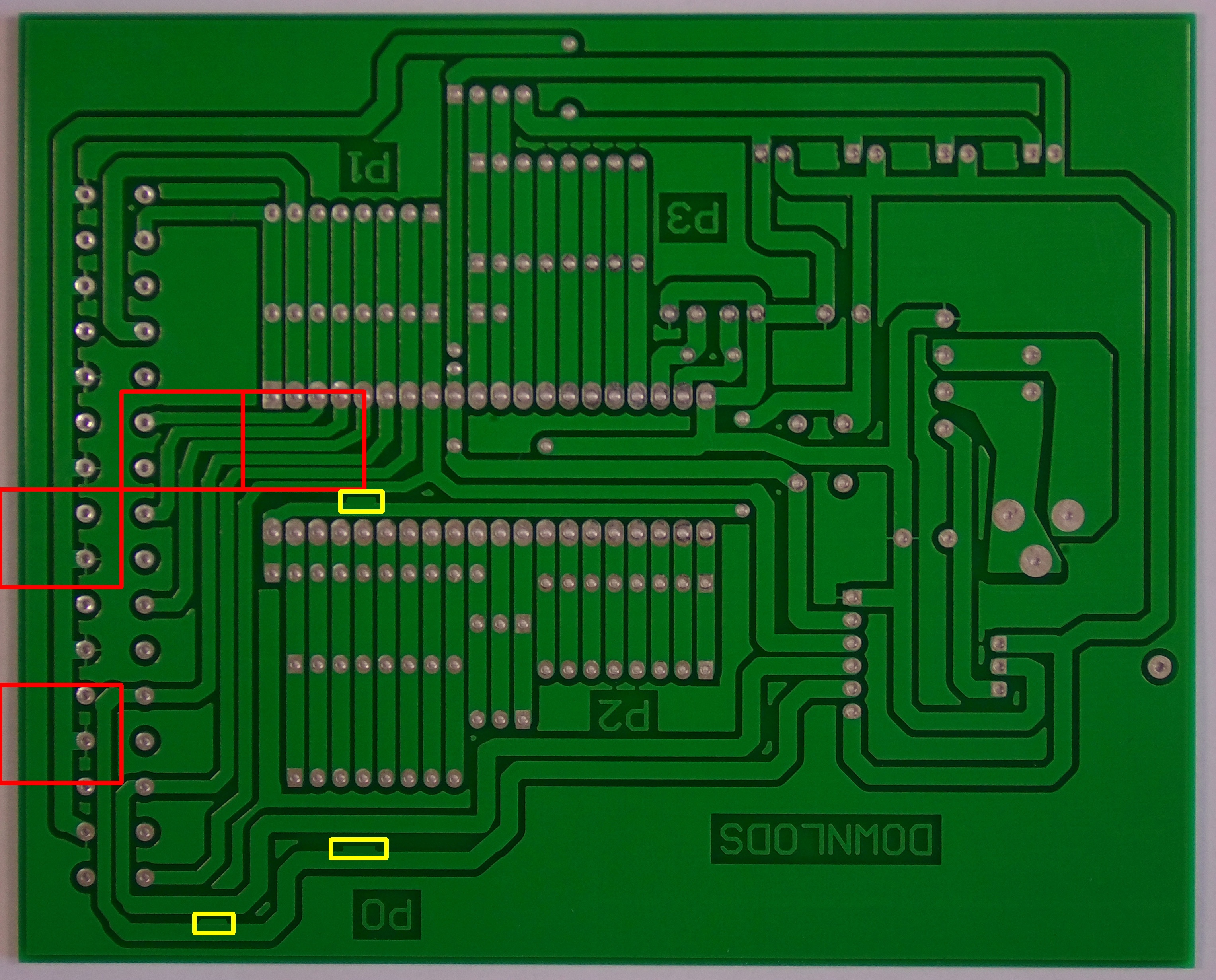}}
    \hfill
    
  \subfloat[EfficientNetV2B0: Open Circuit\label{6c}]{%
        \includegraphics[width=0.49\linewidth]{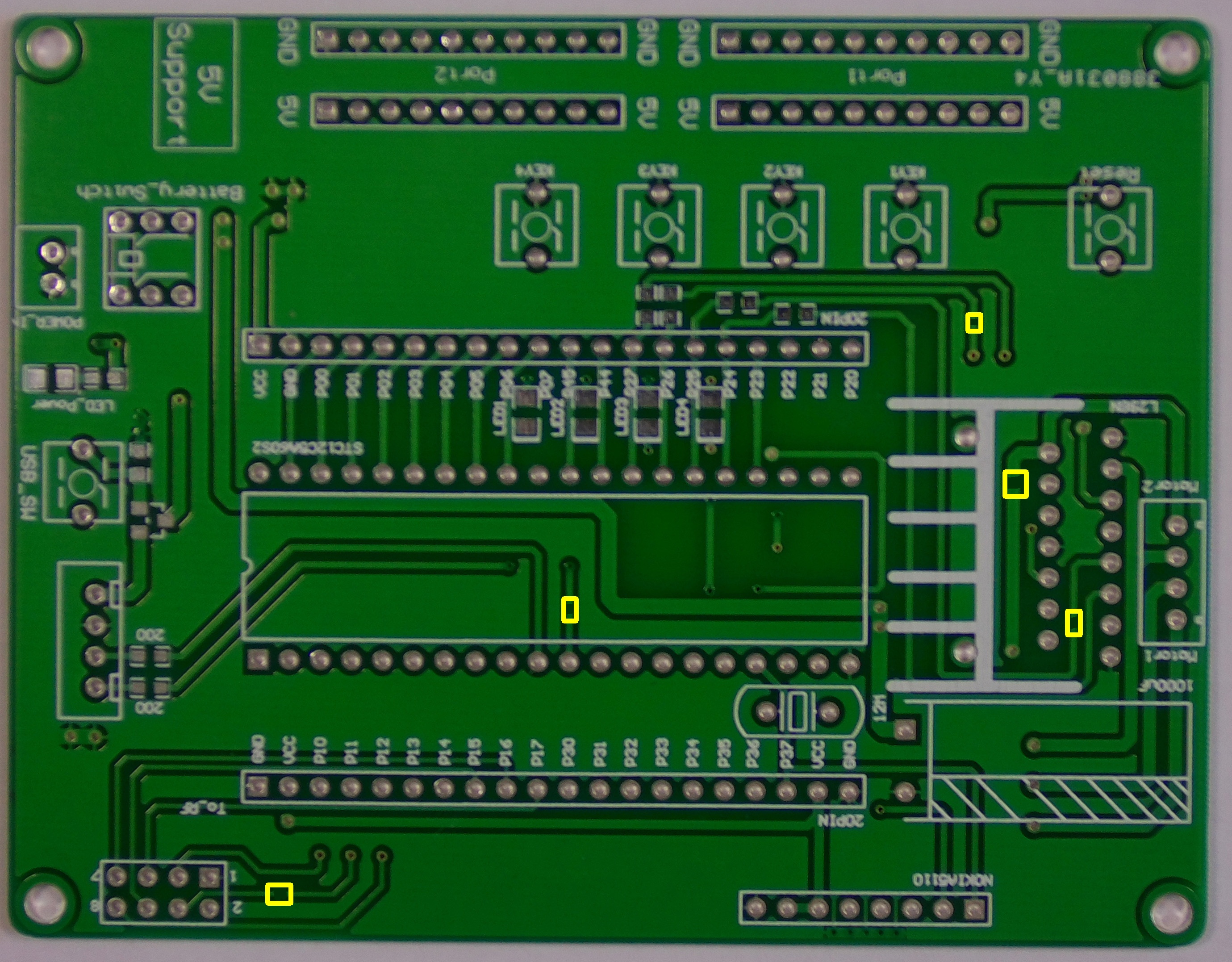}}
    \hfill
  \subfloat[EfficientNetV2B0: Spur\label{6c}]{%
        \includegraphics[width=0.48\linewidth]{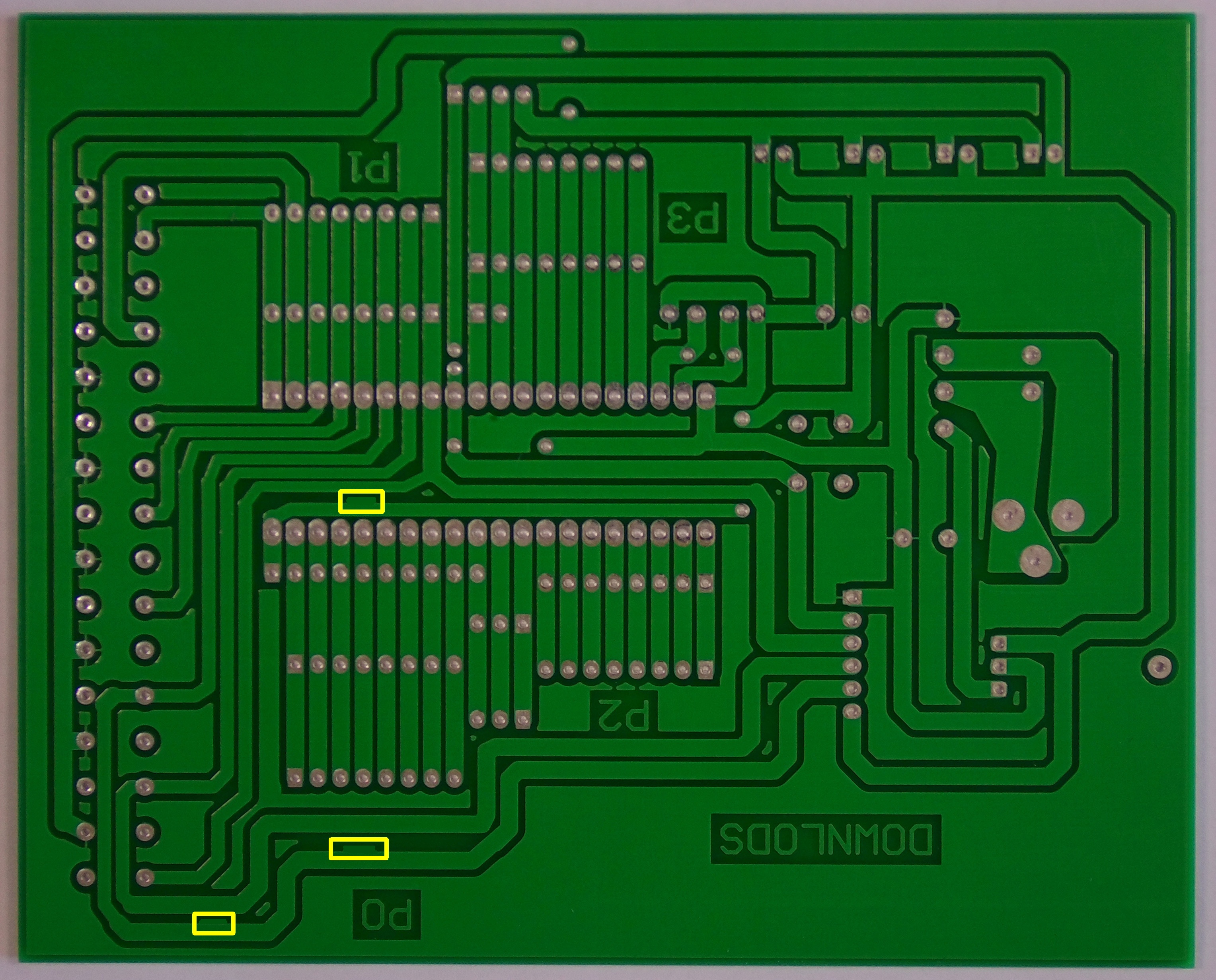}}
    \hfill
    \subfloat[Proposed Model: Open Circuit\label{6a}]{%
       \includegraphics[width=0.49\linewidth]{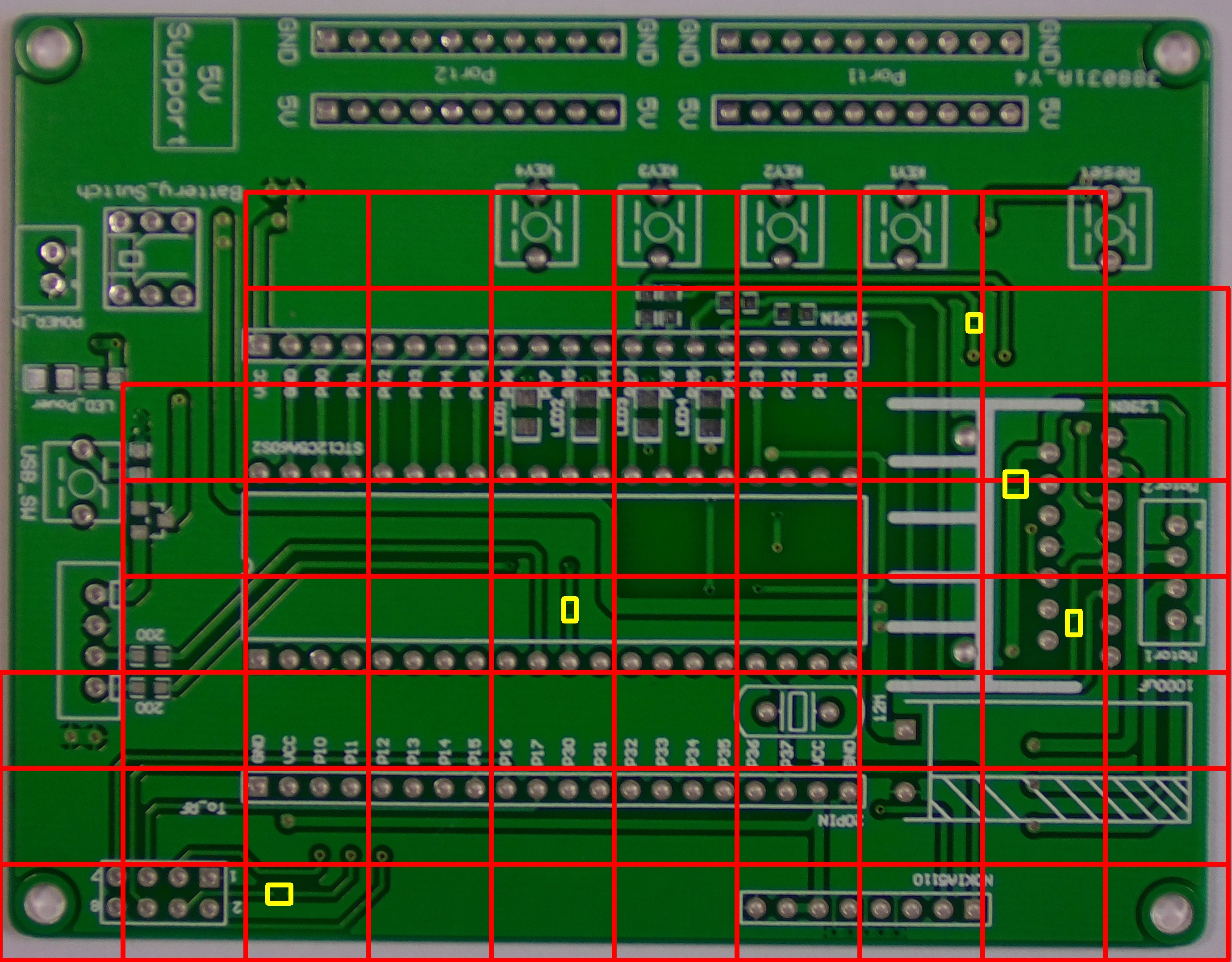}}
    \hfill
    \subfloat[Proposed Model: Spur\label{6b}]{%
       \includegraphics[width=0.48\linewidth]{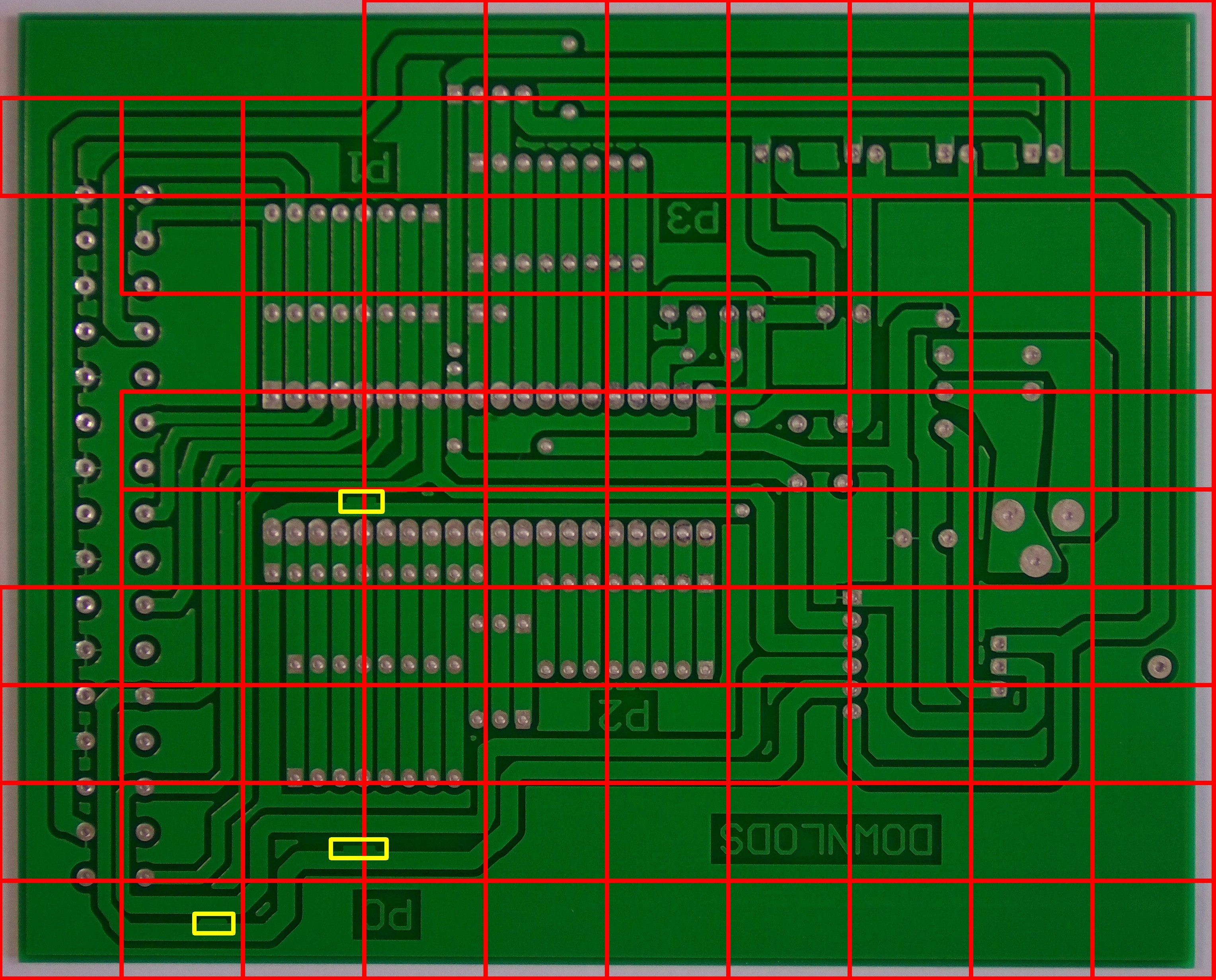}}
    \hfill
  \caption{Comparison of Pre-trained models vs Our Non pre-trained model}
  \label{fig:detection_results_TL}
\end{figure}

\begin{figure}[h!] 
    \centering
  \subfloat[Open Circuit\label{6i}]{%
        \includegraphics[width=0.49\linewidth]{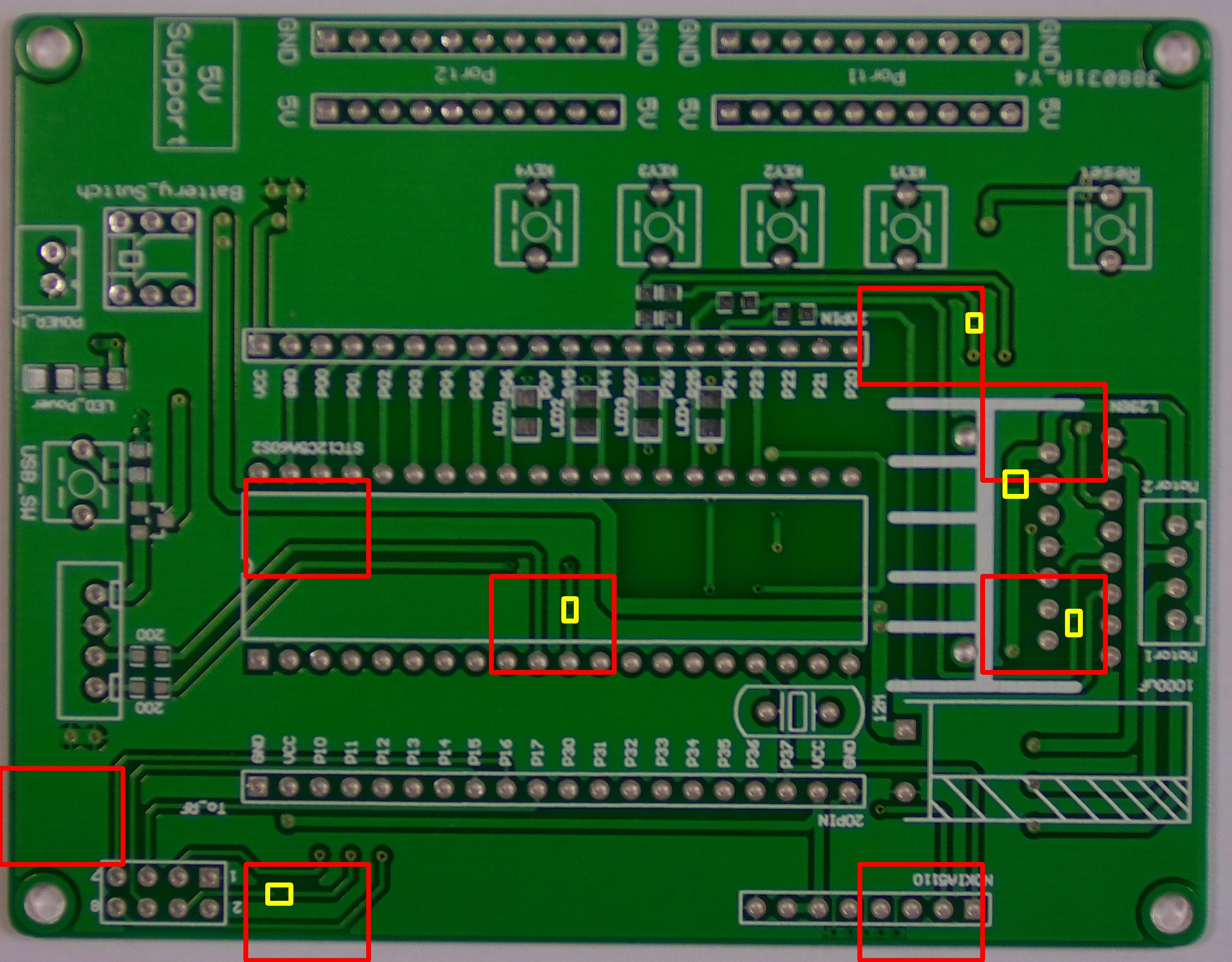}}
     \hfill
     \subfloat[Spur\label{6j}]{%
        \includegraphics[width=0.48\linewidth]{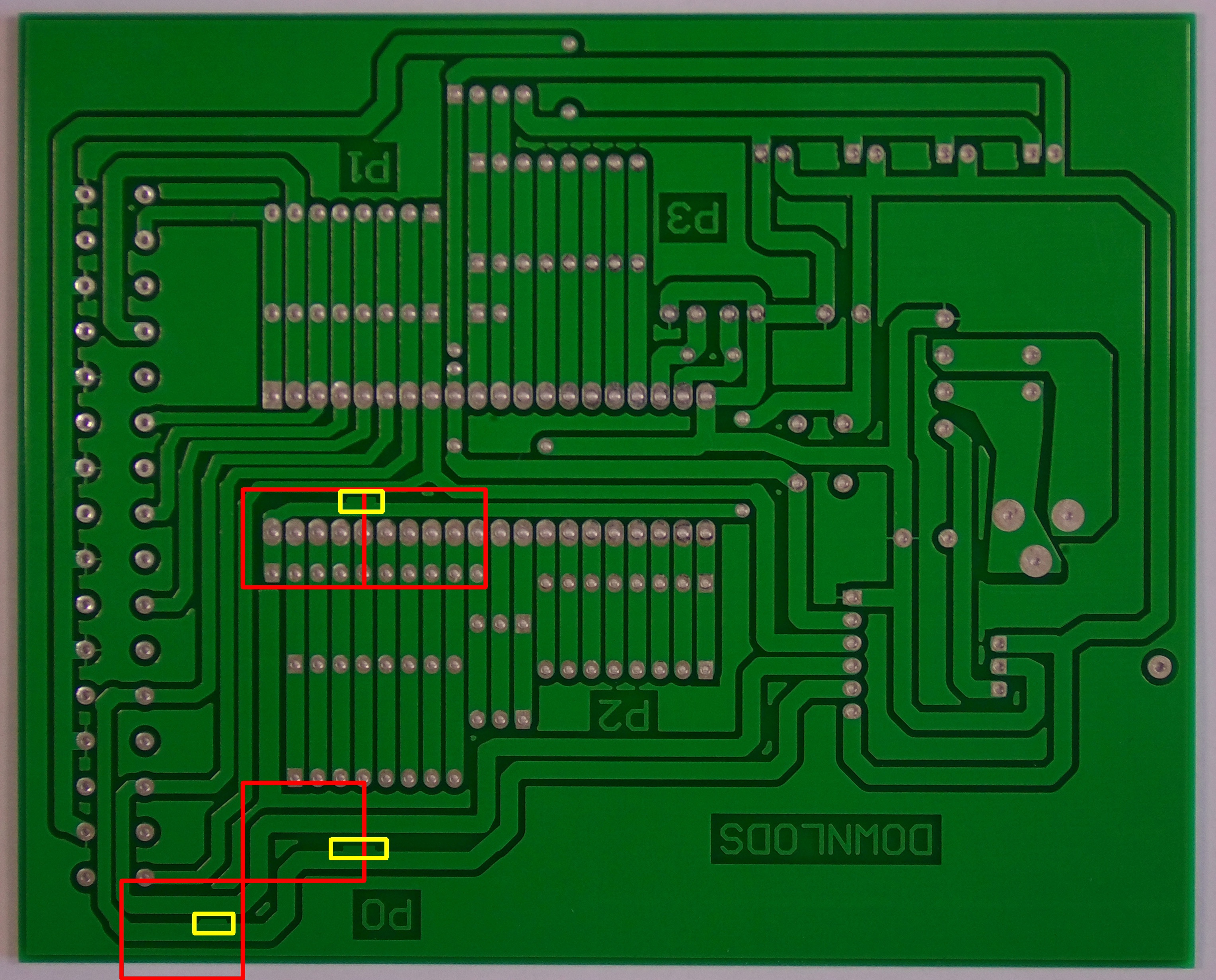}}
     \hfill
  \caption{Benchmark result with Xception (pre-trained on ImageNet) implemented on desktop}
  \label{fig:detection_results_benchmark}
\end{figure}

\begin{figure}[h!] 
    \centering
  \subfloat[Proposed Model: Open Circuit\label{6a}]{%
       \includegraphics[width=0.49\linewidth]{Contents/figures/inf_opencirc_xlite.jpg}}
    \hfill
    \subfloat[Proposed Model: Spur\label{6b}]{%
       \includegraphics[width=0.48\linewidth]{Contents/figures/inf_spur_xlite.jpg}}
    \hfill
      
  \subfloat[EfficientNetV2B0: Open Circuit\label{6e}]{%
        \includegraphics[width=0.49\linewidth]{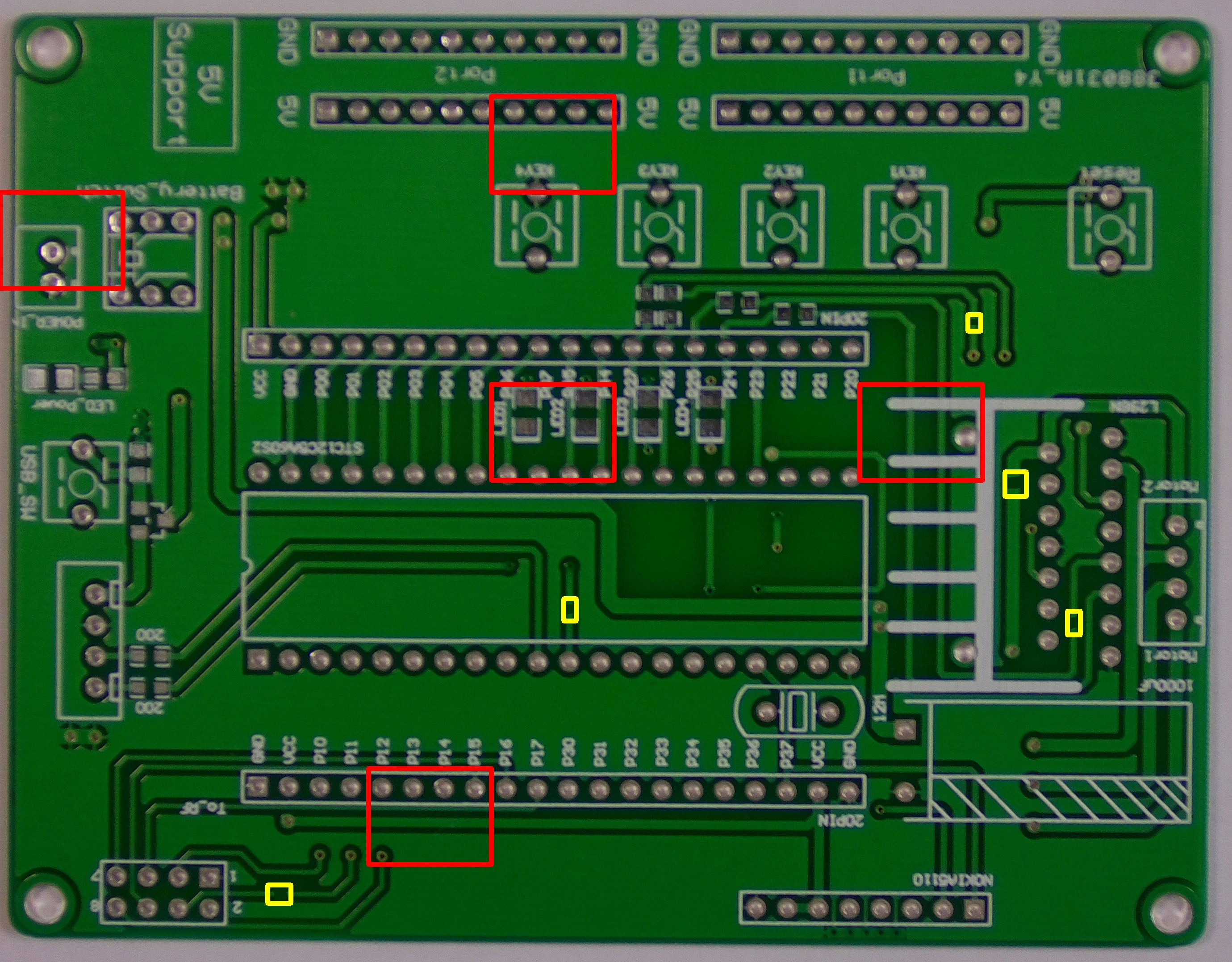}}
    \hfill
    \subfloat[EfficientNetV2B0: Spur\label{6f}]{%
        \includegraphics[width=0.48\linewidth]{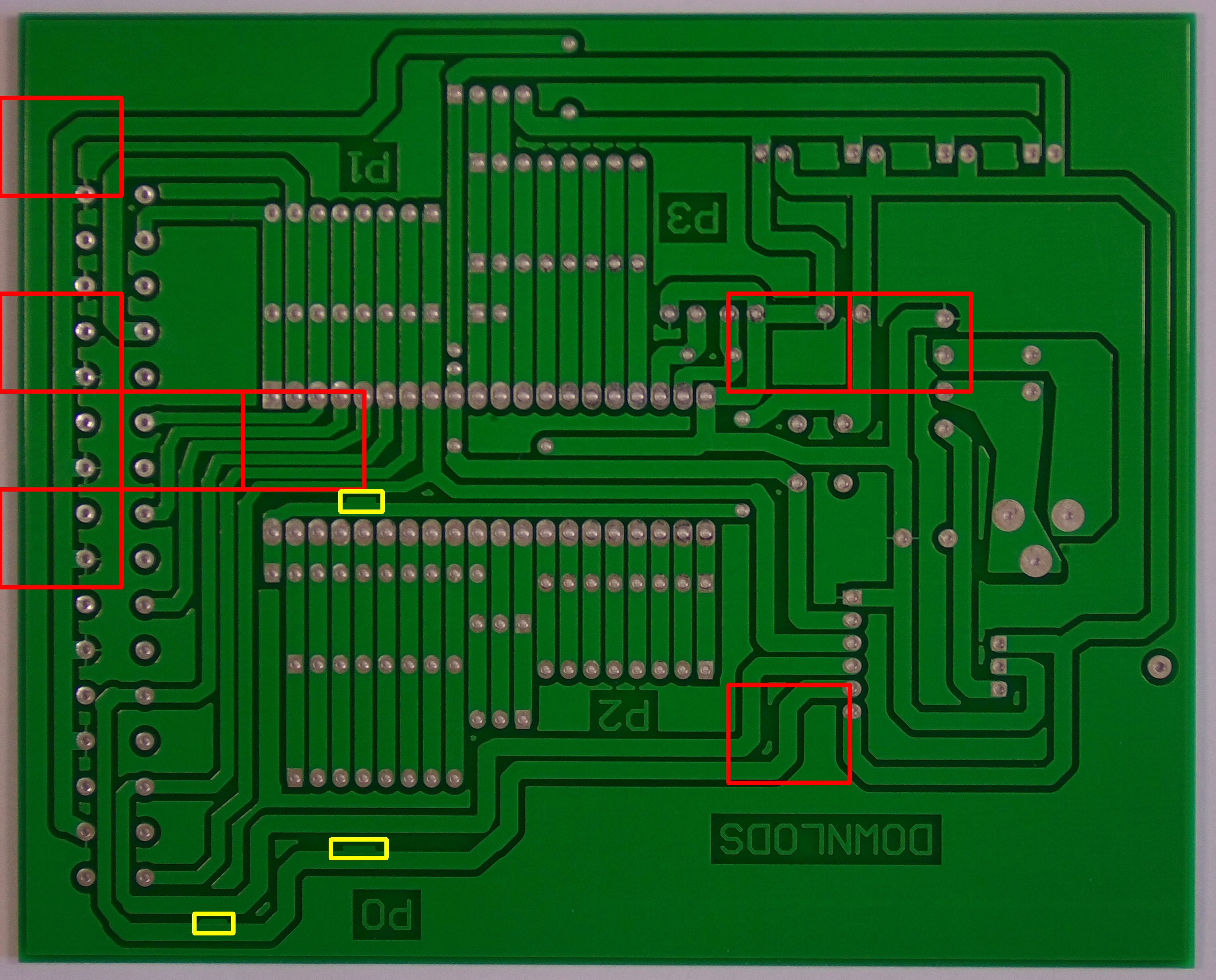}}
    \hfill
    
  \subfloat[MobileNetV2: Open Circuit\label{6c}]{%
        \includegraphics[width=0.49\linewidth]{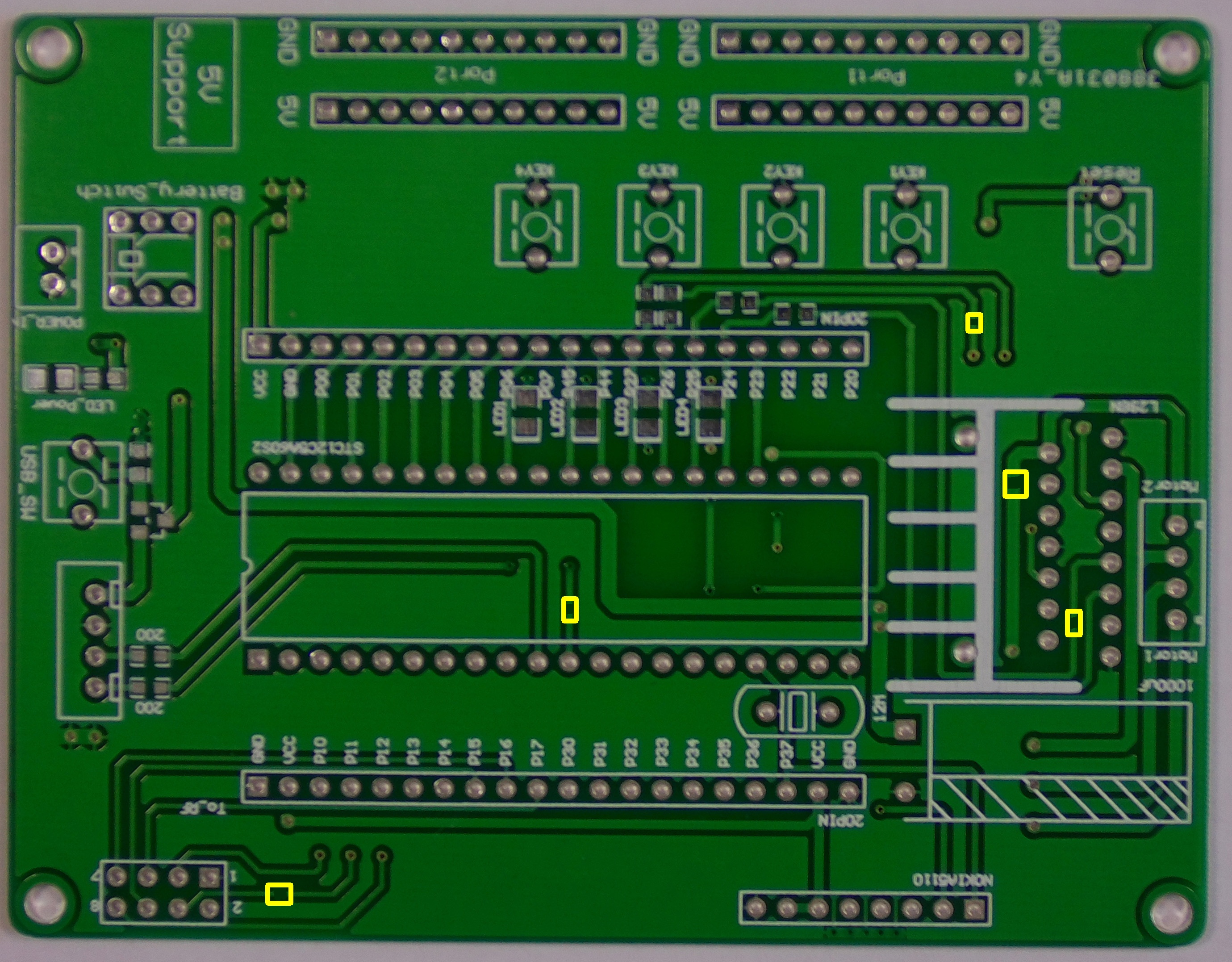}}
    \hfill
    \subfloat[MobileNetV2: Spur\label{6d}]{%
        \includegraphics[width=0.48\linewidth]{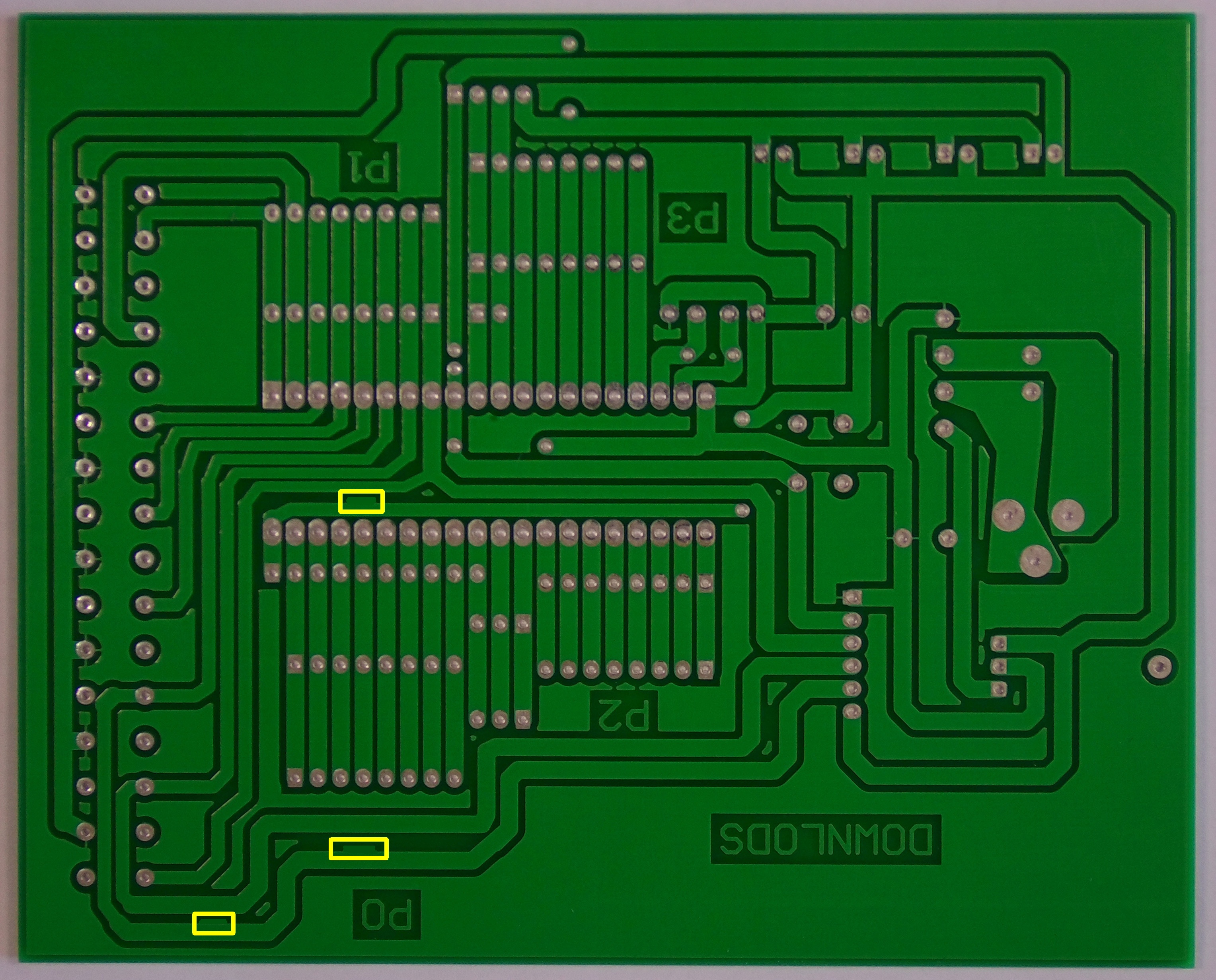}}
    \hfill
    
  \subfloat[MobileViT-XXS: Open Circuit\label{6g}]{%
        \includegraphics[width=0.49\linewidth]{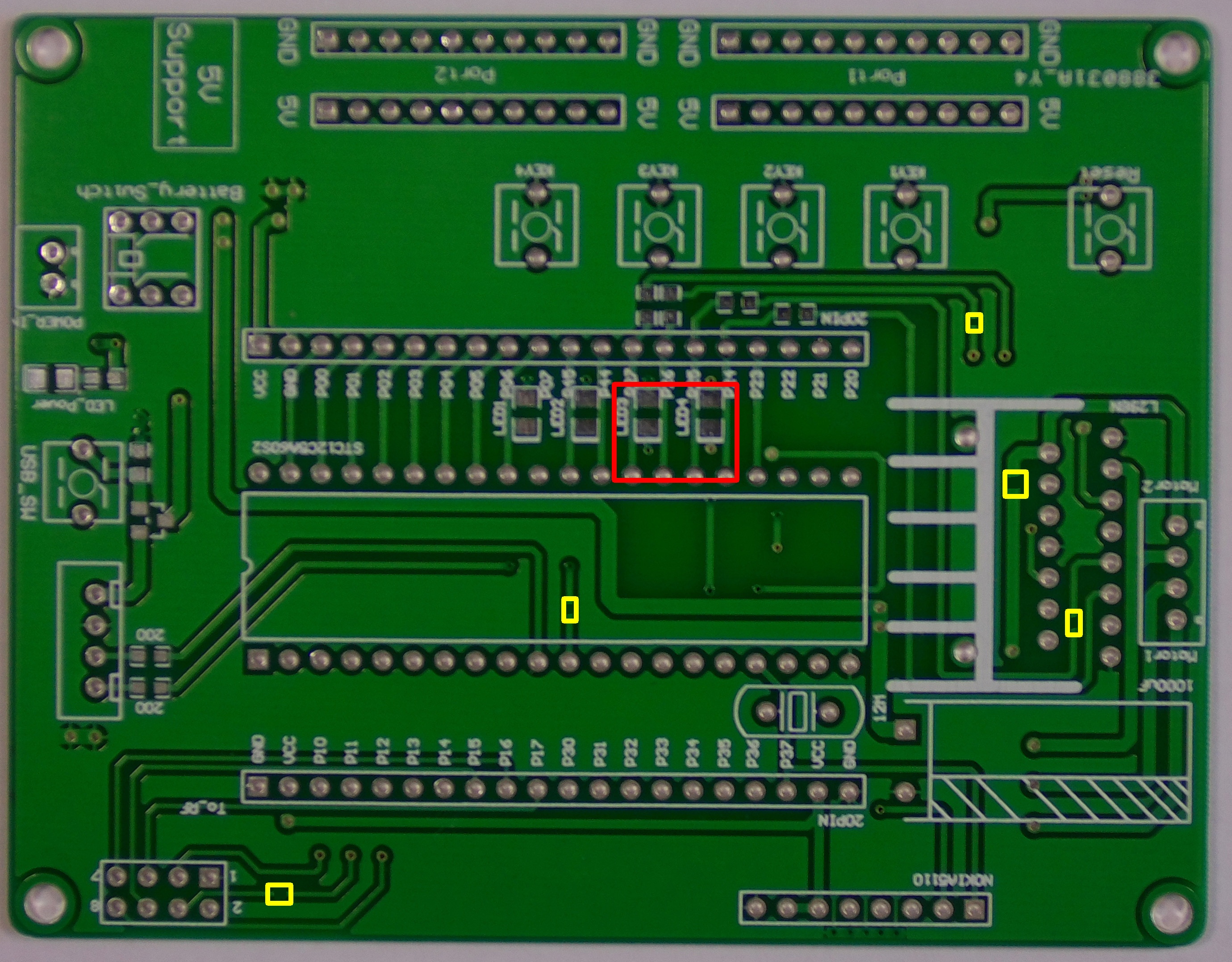}}
    \hfill
    \subfloat[MobileViT-XXS: Spur\label{6h}]{%
        \includegraphics[width=0.48\linewidth]{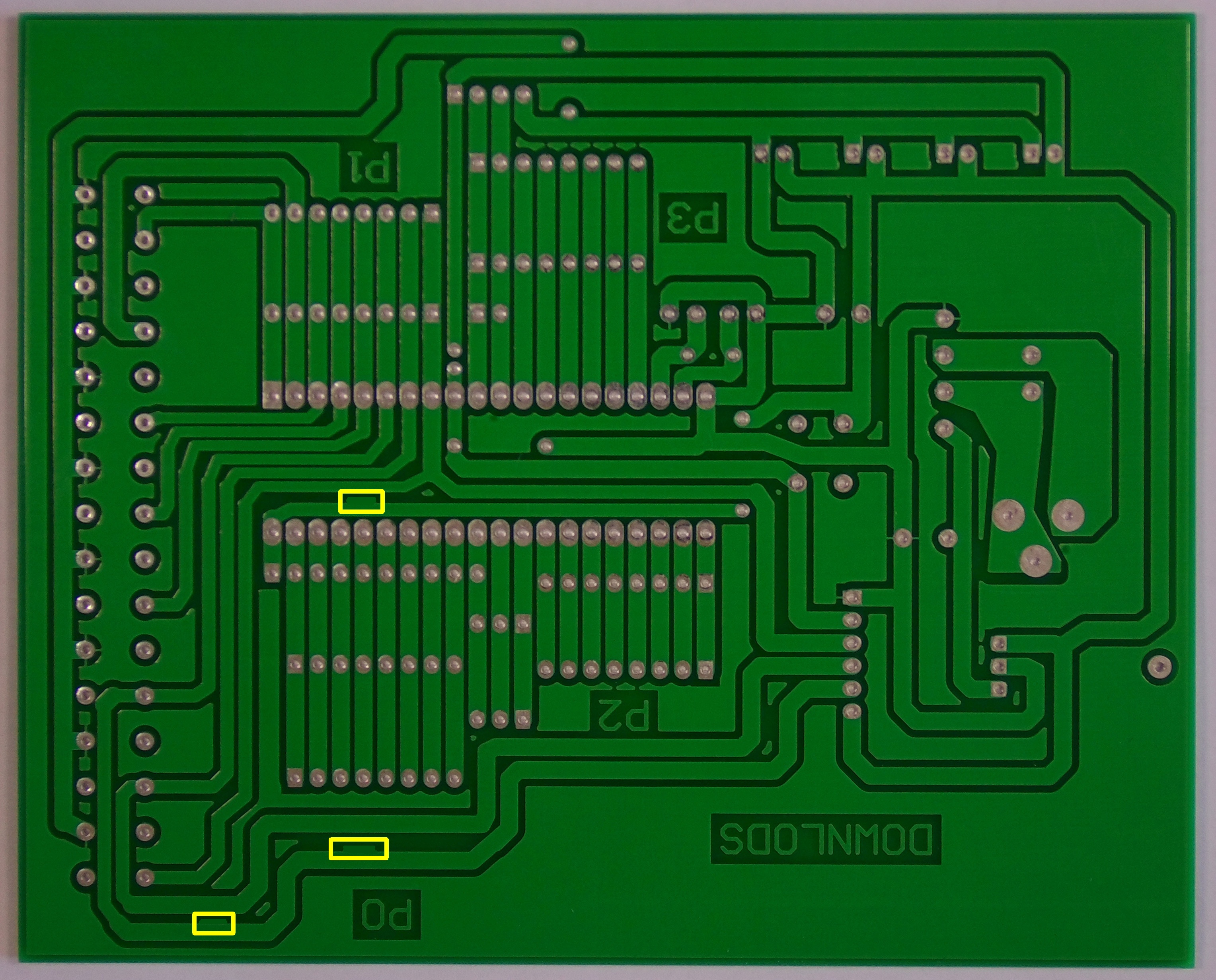}}
    \hfill

  \caption{Result of defect detection on Open Circuit and Spur classes}
  \label{fig:detection_results_OC}
\end{figure}

\end{document}